\documentclass[runningheads]{llncs}
\usepackage[T1]{fontenc}
\usepackage{graphicx}
\usepackage{xcolor}
\usepackage{amsmath}
\usepackage{todonotes}
\usepackage{amssymb}
\usepackage{booktabs}
\usepackage{multirow}
\usepackage{float}
\usepackage{enumitem}
\usepackage{subcaption}
\usepackage{algorithm}
\usepackage[noend]{algpseudocode}
\usepackage{longtable}
\usepackage{adjustbox}
\usepackage{hyperref}
\usepackage{color}

\begin{document}

\title{Linear Ordering Problem: Time for a Change}
\titlerunning{Linear Ordering Problem: Time for a Change}

\author{Fabrizio Fagiolo\inst{1}\orcidID{0009-0003-0390-7855} \and
Marco Baioletti\inst{2}\orcidID{0000-0001-5630-7173} \and
Valentino Santucci\inst{1}\orcidID{0000-0003-1483-7998}}
\authorrunning{F. Fagiolo et al.}
%
\institute{University for Foreigners of Perugia, Perugia 06123, Italy\\
\email{f.fagiolo@phd.unistrapg.it, valentino.santucci@unistrapg.it} \vspace{0.2cm} \and
University of Perugia, Perugia 06123, Italy\\
\email{marco.baioletti@unipg.it}
}

\maketitle              

\begin{abstract}
The Linear Ordering Problem (LOP) is a fundamental combinatorial optimization problem with important applications in areas such as economics, social choice, and machine learning. Its most prominent use is the triangulation of economic input-output tables, which helps identify critical industries in an economy. Most existing algorithms have been evaluated on benchmarks derived from outdated macroeconomic data, which no longer reflect the structure of contemporary economies. Furthermore, LOP instances often exhibit many distinct global optima that can differ substantially from one another, creating challenges for applications that rely on a single solution. To address these limitations, we introduce a novel benchmark suite derived from up-to-date real-world economic data and an algorithmic scheme that leverages state-of-the-art LOP metaheuristics to generate diverse sets of high-quality solutions, together with metrics for assessing both quality and diversity. Experiments were conducted to report results on the proposed benchmark suite under both the traditional single-solution setting and the newly introduced multi-solution scenario.
\keywords{Linear Ordering Problem \and Multimodal Combinatorial Optimization \and Benchmarking \and Permutation Diversity.}
\end{abstract}

\section{Introduction}\label{sec:intro}

The Linear Ordering Problem (LOP) is a fundamental combinatorial optimization problem that has been extensively studied due to its central role in a variety of economically and computationally relevant applications~\cite{marti2022exact}.
Its most prominent application is the triangulation of economic input-output tables~\cite{chiarini2004new,kondo2014triangulation,marti2022exact}, a core task in input-output analysis~\cite{leontief1986input}, the field for which Leontief received the Nobel Prize in 1973.
Beyond economics, the LOP has also found applications in diverse areas, including social choice~\cite{ando2022strong}, sports analytics~\cite{cameron2021linear}, anthropology~\cite{glover1974optimal}, natural language processing and translation~\cite{kim2022chapter,tromble2009learning}, machine learning~\cite{mishra2022linear}, and prompt engineering for generative AI~\cite{11271420}.
Moreover, the LOP has recently attracted attention in studies examining the difficulty and structural characteristics of combinatorial problem instances~\cite{elorza2023characterizing,hernando2020journey,ma2025theoretical,santucci2024optimization} 

The LOP involves finding a permutation that reorders the rows and columns of a given matrix so as to maximize the sum of its upper-triangular part.
When interpreted as the weight matrix of a digraph, the LOP is equivalent to finding a maximum-weight tournament subgraph~\cite{marti2022exact}.
In economic applications, the graph nodes represent industries, while the arcs correspond to inter-industry flows. 
The resulting triangulation provides a structured representation of these relationships and helps to identify critical industries within an economy.

The vast majority of algorithms proposed in the literature have been evaluated using the same benchmark suite, whose most recent version dates back to 2010~\cite{marti2012benchmark}.
This suite includes a mix of artificial and real-world instances, although the latter are based on macroeconomic data that are now between 47 and 67 years old.
In the meantime, financial crises, pandemics, globalization and rapid digitalization of markets have fundamentally reshaped economies, making many of these historical instances less representative of contemporary economies. 
Moreover, from an algorithmic perspective, although the No Free Lunch theorem does not apply to the LOP---since its objective function is not closed under permutation\footnote{This follows immediately from the skew-symmetry property of the LOP: given an instance, the optimal solution is the reverse of the worst solution, the second-best solution is the reverse of the second-worst solution, and so on.}~\cite{droste2002optimization}---, algorithms' performance can still vary significantly across different instances.
Consequently, updated benchmarks are essential both to accurately assess algorithms' performance in contemporary settings and to ensure that algorithmic design reflects the structural characteristics of modern economic data.

To address this gap, in this work we introduce a novel benchmark suite for the LOP based on up-to-date real-world economic data.
Thanks to advances in macroeconomic data collection, the proposed suite, called \texttt{EXIOBASE}, includes larger realistic instances than previous benchmarks, offering a more representative and challenging testbed for modern algorithm evaluation.

Furthermore, recent studies have shown that LOP instances often admit many distinct global optima rather than a single one~\cite{anderson2022fairness,benito2025multiple,cameron2021linear}.
These optimal solutions can differ substantially from one another, posing significant challenges for analyses that rely on a single solution and motivating the need for methods capable of identifying representative and diverse sets of global optima.
Recent approaches mostly rely on exact algorithms, which are only feasible for relatively small instances due to the NP-hardness of the LOP.

To address this limitation, this work introduces an algorithmic scheme designed to leverage state-of-the-art LOP metaheuristics to generate a diverse set of high-quality solutions.
We also propose appropriate metrics for evaluating the quality and diversity of the obtained solutions, providing a rigorous framework for experimental assessment.

The rest of the paper is organized as follows.
Sect.~\ref{sec:lop} formally defines the LOP, reviews state-of-the-art algorithms, summarizes commonly adopted benchmarks, and discusses recent approaches for collecting multiple solutions.
Sect.~\ref{sec:exiobase} introduces the newly proposed real-world economic benchmark suite, while Sect.~\ref{sec:problem} formally defines the Multi-Solution LOP (MS-LOP). 
The algorithmic scheme for generating multiple high-quality solutions is presented in Sect.~\ref{sec:algs}, and experimental results are reported and analyzed in Sect.~\ref{sec:exp}.
Finally, Sect.~\ref{sec:concl} concludes the paper by also outlining directions for future research.


\section{Linear Ordering Problem}\label{sec:lop}

An LOP instance is represented by a square matrix $\mathbf{B} = [ b_{ij} ]_{n \times n}$.
Denoting the set of all permutations of $[n] = \{1,2,\ldots,n\}$ as $\mathbb{S}_n$, the goal is to find an optimal permutation $\sigma^* \in \mathbb{S}_n$ that maximizes the objective function
\begin{equation}
	f(\sigma) = \sum_{i=1}^{n-1} \sum_{j=i+1}^n b_{\sigma(i),\sigma(j)}
    .
	\label{eq:objfun}
\end{equation}


\subsubsection{Algorithms for the LOP.}\label{sec:lop_algs}

The LOP was proven NP-hard in the seminal work of Garey and Johnson~\cite{garey1979computers}, and was later shown to be APX-complete~\cite{mishra2004approximability}.
Consequently, many metaheuristics have been proposed over the past decades (e.g.,~\cite{ali2024comprehensive,baioletti2015linear,baioletti2017new,campos2001experimental,ceberio2023model,garcia2006variable,laguna1999intensification,marti2022exact,qian2020block,sakuraba2010efficient}).

To the best of our knowledge, the current state-of-the-art methods for the LOP are CD-RVNS~\cite{santucci2020using} and MA-EDM~\cite{lugo2022diversity}.


CD-RVNS, introduced in~\cite{santucci2020using}, alternates between a variable neighborhood search phase and a destruction-construction phase.
The variable neighborhood search identifies a local optimum under both insertion and interchange neighborhoods,
using the restricted neighborhood technique 
for the LOP
proposed in~\cite{ceberio2015linear}.
The destruction and construction operators partially disrupt the local optimum and then reconstruct a new solution by exploiting a novel representation specifically designed for the LOP.
These latter procedures employ both intensification and diversification mechanisms, which appear to be the reason for the superior performance of CD-RVNS with respect to previous methods.

MA-EDM was introduced in~\cite{lugo2022diversity} and its core idea is to maintain a population of local optima, under the insertion neighborhood, that is both diverse and of high quality.
To this end, it employs a population selection procedure based on the permutation deviation distance, previously introduced in~\cite{sevaux2005permutation}, which can be computed in linear time and is particularly well-suited for the LOP.
Unlike 
\mbox{CD-RVNS}, the local search incorporated in MA-EDM does not use the restricted neighborhood, but relies on a classical acceleration technique that allows the evaluation of all the neighbors in $\Theta(n^2)$ time~\cite{schiavinotto2007review}.

\subsubsection{Benchmark suites for the LOP.}\label{sec:lop_bench}
\texttt{LOLIB} is the most extensively studied benchmark suite for the LOP.
It was compiled in 2010 and presented in~\cite{marti2012benchmark}.
The suite originally comprised 485 instances grouped into eight distinct sets and was later extended in~\cite{ceberio2015linear} with the addition of the \texttt{xLOLIB2} instance set\footnote{\texttt{LOLIB} and \texttt{xLOLIB2} are available at \url{https://www.uv.es/rmarti/paper/lop.html} and \url{https://github.com/sgpceurj/EJOR2015}, respectively.}.
The resulting suite, now consisting of nine sets and 685 instances, has become the de facto standard in LOP research.
All the state-of-the-art algorithms discussed previously~\cite{santucci2020using,lugo2022diversity} have been evaluated on these nine instance sets, which are listed along with their main characteristics in Tab.~\ref{tab:bench}.
Most of the instances are artificial, while only a few are derived from real-world data.

\begin{table}[t]\scriptsize
    \setlength{\tabcolsep}{0.2cm}
    \centering
    \caption{Main characteristics of the benchmark instance sets for the LOP.}
    \resizebox{\textwidth}{!}{
    \begin{tabular}{lccccl}
        \toprule
        \textbf{Instance Set} & \textbf{\#Instances} & \textbf{Size} & \textbf{Normal Form} & \textbf{\#Optima} & \textbf{Type} \\
        \midrule
        \texttt{IO}~\cite{grotschel1984cutting}       & \phantom{0}50 & 44-79               & Yes & 50           & Economic IO tables (1959--1979) \\
        \texttt{SGB}~\cite{laguna1999intensification} & \phantom{0}25 & 75                  & Yes & 25           & Random uniform from large range \\
        \texttt{RandA1}~\cite{campos2001experimental} & 100           & 100-500             & Yes & \phantom{0}0 & Random uniform from small range\\
        \texttt{RandA2}~\cite{campos2001experimental} & \phantom{0}75 & 100-200             & Yes & \phantom{0}0 & From random permutations \\
        \texttt{RandB}~\cite{marti2012benchmark}      & \phantom{0}90 & 40-50               & Yes & 71           & Random asymmetric \\
        \texttt{MB}~\cite{mitchell2000solving}        & \phantom{0}30 & 100-250             & Yes & 30           & Random asymmetric + zeroed entries \\
        \texttt{xLOLIB}~\cite{schiavinotto2004linear} & \phantom{0}78 & 150-250             & Yes & \phantom{0}0 & Oversampled \texttt{IO} instances \\
        \texttt{Spec}~\cite{christof1996combinatorial,goemans1996strongest,marti2012benchmark} & \phantom{0}37 & \phantom{0}11-452   & Yes & 30           & Artificial, subsampled \texttt{IO}, tennis results \\
        \texttt{xLOLIB2}~\cite{ceberio2015linear}     & 200           & \phantom{0}300-1000 & No  & \phantom{0}0 & Oversampled \texttt{IO} instances \\
        \bottomrule
    \end{tabular}
    }
    \label{tab:bench}
\end{table}

The only proper real-world instances are the national input-output tables in the \texttt{IO} set, first used in the context of the LOP in~\cite{grotschel1984cutting} and based on data from Eurostat and other statistical agencies.
These IO tables span the years \mbox{1959--1979} and include 50 instances: 31 with $n=44$, 4 with $n=50$, 11 with $n=56$, 3 with $n=60$, and one (\texttt{usa79}) with $n=79$.
Schiavinotto and St\"utzle~\cite{schiavinotto2004linear} then generated the \texttt{xLOLIB} instances by oversampling 39 \texttt{IO} instances: for each, they created two larger instances with $n=150$ and $n=250$ by uniformly sampling entries from the corresponding IO table.
They describe these \texttt{xLOLIB} instances as ``real-life like'' problems.
The same procedure was later used by Ceberio~et~al.~\cite{ceberio2015linear} to generate 
the \texttt{xLOLIB2}
instances, 50 for each size $n \in \{300, 500, 750, 1000\}$.
Finally, the \texttt{Spec} set---comprising both artificial and real-world instances---contains 15 economic-like matrices generated in~\cite{christof1996combinatorial} by subsampling the \texttt{usa79} instance from the \texttt{IO} set, as well as 8 preference counting matrices derived from the results of ATP tennis tournaments held in 1993–-1994.

All remaining instances are artificially generated.
The entries of \texttt{SGB} and \texttt{RandA1} instances are sampled uniformly at random from 
$[0,250\,000]$ 
and $[0,100]$, respectively.
\texttt{RandA2} instances are preference counting matrices constructed from a set of randomly generated permutations.
\texttt{RandB} and \texttt{MB} instances are generated with upper and lower triangular parts drawn from uniform distributions over intervals of different length, with \texttt{MB} instances further processed by zeroing-out a certain percentage of entries.
Finally, the 14 fully artificial instances in \texttt{Spec} include 6 random binary matrices and 8 Paley graph adjacency matrices.

The number of known optima reported in Tab.~\ref{tab:bench} (proven via integer programming) is taken from~\cite{lugo2022diversity}.
In general, the unsolved instances are the larger ones, with a few exceptions: \texttt{MB} instances are easier due to the zeroing-out process, and \texttt{RandA1} instances appear more difficult than \texttt{SGB} instances because of their different variability.
Indeed, both sparsity (the percentage of zeros) and entry variability are known to influence instance difficulty~\cite{schiavinotto2004linear}.

All benchmark instances have non-negative integer entries.
Moreover, all instances except those in \texttt{xLOLIB2} are in normal form, i.e., \mbox{$\min\{b_{ij},b_{ji}\}=0$} for all 
$i,j \in [n]$.
As noted in~\cite{marti2012benchmark}, every LOP instance can be normalized without altering the ranking over the solutions induced by the objective function.



\subsubsection{Multiple high-quality solutions for the LOP.}\label{sec:lop_multi}

Recently, there has been a growing interest in the observation that LOP instances often have multiple distinct global optima rather than a single one. 
Benito-Marimom et al.~\cite{benito2025multiple} showed, using integer programming, that: all \texttt{IO} instances have more than one global optimum, only 7 out of 50 instances had fewer than 500 optima, while 21 exhibited more than $500\,000$ distinct optima.
Although these solutions share the same optimal objective value, they can differ substantially in their genotypic representations.
In particular, \cite{anderson2022fairness} showed that some items may vary considerably in both their positions and relative ordering across the different optimal solutions.

Therefore, novel algorithms aimed at identifying representative sets of multiple global optima for the LOP have begun to emerge.
Cameron et al.~\cite{cameron2021linear} proposed an integer programming model to identify two global optima that are maximally distant according to the Kendall’s-$\tau$ distance~\cite{schiavinotto2007review}.
Building on this idea, Anderson et al.~\cite{anderson2022fairness} introduced a family of integer programming models to: compute a summary---referred to as the ``centroid''---of item positions across all global optima, identify pairs of optima closest to and farthest from this centroid, as well as the most similar and most dissimilar pairs of optima.
In all cases, the Kendall’s-$\tau$ distance is used to measure (dis)similarity, as it is particularly well suited to the LOP.


More recently, \cite{benito2025multiple} extended the model of~\cite{cameron2021linear} to generate sets of more than two global optima that are maximally distant from one another.
To this end, they considered two distinct objectives: maximizing the sum of all pairwise distances and maximizing the minimum distance between any two solutions.
The latter helps to prevent multiple copies of the same optimum and ensures better coverage of the entire set of global optima.
The same authors also noted that the integer programming approaches are limited to small/medium-sized instances and therefore proposed a two-phase heuristic method based on a Variable Neighborhood Search (VNS) scheme using insertion neighborhoods with an increasing number of moves.
In the first phase, VNS finds the best possible objective value (with no theoretical guarantee), while in the second phase, a separate VNS collects $m$ solutions attaining this value into a representative set that tries to maximize the sum of pairwise Kendall's-$\tau$ distances.
The authors explicitly note that this second-phase VNS ignores any solutions with an objective value higher than that found in the first phase.

Beyond the LOP, two related areas of research can be identified in the evolutionary computation field: multimodal optimization and evolutionary diversity optimization.
Multimodal optimization~\cite{preuss2015multimodal,preuss2021multimodal} deals with the task of approximating the set of local optima
whose quality is higher than a specified threshold.
In this context, the diversity of solutions is primarily considered to enhance the search exploration, rather than being an explicit optimization goal in its own~\cite{do2022analysis}.
In contrast, evolutionary diversity optimization, introduced in \cite{ulrich2011maximizing}, aims to compute a set of solutions whose quality is higher than a given threshold and that are maximally diverse from one another.
While multimodal optimization is mainly applied to continuous search spaces (see the recent GECCO competition~\cite{ahrari2024experimental}), evolutionary diversity optimization has mostly been studied from a theoretical perspective (e.g.,~\cite{bossek2021evolutionary,do2022analysis}).
Neither has been applied to the LOP.
Moreover, most of the works in these research lines assume that the optima or their objective values are known in advance. 


\section{The \texttt{EXIOBASE} Benchmark Suite}\label{sec:exiobase}

To address the limitations of existing benchmark suites for the LOP, we introduce a novel suite, \texttt{EXIOBASE}, derived from real-world data in the homonymous input-output (IO) database introduced by Stadler~et~al.~\cite{stadler2018exiobase} for economical and sustainability analyses.
The proposed benchmarks are publicly available at \url{https://doi.org/10.5281/zenodo.20554376}.

The rest of this section outlines the structure of the original IO tables, describes the construction of our benchmark suite, and presents comparative statistics with classical benchmarks.

\subsubsection{The original IO database.}
EXIOBASE~\cite{stadler2018exiobase} is a widely used global multi-regional input-output 
database that links economic transactions between countries and industries with environmental and social indicators; it is standardly applied in footprint and sustainability assessments.
At the time of writing, the most recent release is version~3.9.6, available at~\cite{stadler_2025_15689391}, which provides annual data from 1995 to 2022.
While environmental extensions are available, the present study focuses exclusively on the economic component containing the IO tables.
For each year, four primary economic data structures are provided: the multi-regional transaction matrix $\mathbf{Z}$, the final demand matrix $\mathbf{Y}$, the total output vector $\mathbf{x}$, and the structural matrix $\mathbf{A}$.

Data are available in both industry-by-industry and product-by-product formats.
Here, we focus on the product-by-product format.
In this configuration, the rows and columns of $\mathbf{Z} = [z_{ij}]_{s \times s}$ correspond to specific region-product pairs.
Each entry $z_{ij}$ represents the monetary value of the output from \mbox{region-product~$i$} used as intermediate input by region-product~$j$, expressed in millions of euros.
The order of $\mathbf{Z}$ is $s = s_{\mathrm{R}} \cdot s_{\mathrm{P}}$, with $s_{\mathrm{R}}=49$ regions (44~countries and 5~rest-of-the-world regions) and $s_{\mathrm{P}}=200$ products.
The matrix $\mathbf{Y} \in \mathbb{R}^{s \times f}$ represents final demand, with rows corresponding to the $s$ region-product pairs and $f = s_\mathrm{R} \cdot s_\mathrm{F}$ columns representing the $s_{\mathrm{F}} = 7$ standard final demand categories repeated for each of the $s_{\mathrm{R}}$ regions.
The output vector $\mathbf{x} \in \mathbb{R}^s$ contains the total production of each sector and is computed as $\mathbf{x} = \mathbf{Z} \mathbf{1}_s + \mathbf{Y} \mathbf{1}_f$, where $\mathbf{1}_s$ and $\mathbf{1}_f$ are all-ones vectors of length $s$ and $f$, respectively.
Finally, the structural matrix $\mathbf{A} = [a_{ij}]_{s \times s}$ is a normalized version of $\mathbf{Z}$, computed as $\mathbf{A} = \mathbf{Z} \, \mathrm{diag}(\mathbf{x})^{-1}$, where each entry $a_{ij} \ge 0$ is a \textit{technical coefficient} representing the amount of output from \mbox{region-product $i$} required to produce one unit of region-product $j$.
These coefficients are dimensionless ratios of monetary values, and $\mathbf{A}$, like $\mathbf{Z}$, is a $9800 \times 9800$ 
matrix
with the following block structure:
\begin{equation}
    \mathbf{A} =
        \begin{pmatrix}
            \mathbf{A}_{1,1} & \mathbf{A}_{1,2} & \cdots & \mathbf{A}_{1,s_\mathrm{R}} \\
            \mathbf{A}_{2,1} & \mathbf{A}_{2,2} & \cdots & \mathbf{A}_{2,s_\mathrm{R}} \\
            \vdots  & \vdots  & \ddots & \vdots  \\
            \mathbf{A}_{s_\mathrm{R},1} & \mathbf{A}_{s_\mathrm{R},2} & \cdots & \mathbf{A}_{s_\mathrm{R},s_\mathrm{R}}
        \end{pmatrix}
    ,
    \label{eq:exiobase}
\end{equation}
where each block $\mathbf{A}_{kl}$ is a $s_{\mathrm{P}} \times s_{\mathrm{P}}$ matrix, with diagonal blocks $\mathbf{A}_{kk}$ representing inter-industry flows within region $k$, and the off-diagonal blocks $\mathbf{A}_{kl}$ ($k \neq l$) describing inter-regional trade.

\subsubsection{Construction of the benchmark suite.}
Building on the data described above, we introduce the \texttt{EXIOBASE} benchmark suite for the LOP, consisting of four distinct instance sets: \texttt{pxp}, \texttt{rxr}, \texttt{isic}, and \texttt{os300}.
These sets are designed to reflect different levels of aggregation and capture a range of contemporary global economic interdependencies.

The set \texttt{pxp} (product-by-product) consists of 49 instances derived from the domestic structural matrices for the year 2022.
Each instance is obtained from the corresponding diagonal block $\mathbf{A}_{kk}$ in Eq.~\eqref{eq:exiobase}, representing the inter-industry transactions for region $k \in \{1, 2, \dots, 49\}$.
Formally, given a \mbox{$200 \times 200$} domestic matrix $\mathbf{A}$, a corresponding LOP instance $\mathbf{B}$ is constructed in two steps.
First, an intermediate matrix $\mathbf{B}'$ is computed as
$\mathbf{B}' = \lfloor c \mathbf{A} - \min(c \mathbf{A}, c \mathbf{A}^T) \rceil$, 
where $c=10^5$ is a scaling factor, both $\mathrm{min}$ and $\lfloor \cdot \rceil$ are applied element-wise, and $\lfloor \cdot \rceil$ denotes the rounding to the nearest integer.
This transformation ensures that instances have integer entries within $[0,10^5]$ and are in normal form, as defined in~\cite{marti2012benchmark}.
Second, the final LOP instance $\mathbf{B}$ is obtained from $\mathbf{B}'$ by removing all products $i$ for which both the $i$-th row and $i$-th column are entirely zero. 
This filtering step is essential, since such ``null products'' can be placed arbitrarily within a permutation without affecting its LOP objective value; consequently, $k$ null products would artificially increase the number of global optima by a factor of $\binom{n}{k} = \Theta(n^k)$.
As a result, the instances in \texttt{pxp} have varying sizes, ranging from 89 to 176, with a median size of 149.

The set \texttt{rxr} (region-by-region) consists of 28 instances, one for each year from 1995 to 2022, obtained by first aggregating the technical coefficients into $49 \times 49$ matrices representing inter-regional trade and then applying the same scaling, rounding, and normalization steps described above.
In this case, all instances have size $n=49$.

The set \texttt{isic} comprises 49 matrices generated by aggregating the 2022 data according to the International Standard Industrial Classification (ISIC).
This aggregation groups the 200 original products into the first ten ISIC categories (e.g., agriculture, manufacturing, and services).
Scaling, rounding and normalization steps are applied here as well.
All instances therefore have size $n=10$ and can be exhaustively evaluated, making them particularly useful in scenarios where exact global optima are required (e.g.,~\cite{ma2025theoretical}).

Finally, the set \texttt{os300} consists of 49 real-life-like instances of size $n=300$.
Each is created by oversampling the corresponding \texttt{pxp} instance using the technique used in~\cite{schiavinotto2004linear}~and~\cite{ceberio2015linear}, followed by normalization of the resulting matrix.

\subsubsection{Statistical analysis of the instances.}
Tab.~\ref{tab:bench_stats} compares the main structural statistics of the \texttt{EXIOBASE} and \texttt{LOLIB} benchmark suites.
In particular, we report the same measures used in~\cite{schiavinotto2004linear}: sparsity (percentage of zeros), the coefficient of variation (ratio of standard deviation to mean), and the skewness index.
Since the instances are in normal form, these statistics were computed only on the maximum of each pair $(b_{ij},b_{ji})$ for all $1 \le i < j \le n$.

\begin{table}[t]\scriptsize
\setlength{\tabcolsep}{0.2cm}
\centering
\caption{Structural statistics for the \texttt{EXIOBASE} and \texttt{LOLIB} benchmark suites.}
\label{tab:bench_stats}
\resizebox{\textwidth}{!}{
\begin{tabular}{llcccccc}
\toprule
 &  & \multicolumn{2}{c}{\textbf{Sparsity}} & \multicolumn{2}{c}{\textbf{Variation Coeff.}} & \multicolumn{2}{c}{\textbf{Skewness}} \\
\textbf{Suite} & \textbf{Set} & Median & [Min,Max] & Median & [Min,Max] & Median & [Min,Max] \\
\midrule
\multirow[t]{4}{*}{\texttt{EXIOBASE}} & \texttt{isic} & $0.00$ & $[0.00, 0.04]$ & $1.23$ & $[0.89, 2.86]$ & $2.14$ & $[0.74, 5.88]$ \\
 & \texttt{rxr} & $0.00$ & $[0.00, 0.01]$ & $1.85$ & $[1.77, 2.30]$ & $4.67$ & $[4.07, 10.02]$ \\
 & \texttt{pxp} & $0.17$ & $[0.03, 0.45]$ & $5.19$ & $[3.03, 8.26]$ & $16.01$ & $[7.88, 27.17]$ \\
 & \texttt{os300} & $0.01$ & $[0.00, 0.02]$ & $3.62$ & $[2.20, 6.20]$ & $10.14$ & $[5.83, 16.69]$ \\
\cline{2-8}
 & \textbf{Overall} & $0.01$ & $[0.00, 0.45]$ & $2.92$ & $[0.89, 8.26]$ & $8.70$ & $[0.74, 27.17]$ \\
\cline{1-8}
\multirow[t]{8}{*}{\texttt{LOLIB}} & \texttt{IO} & $0.18$ & $[0.05, 0.71]$ & $3.74$ & $[2.59, 9.22]$ & $10.54$ & $[4.95, 23.67]$ \\
 & \texttt{SGB} & $0.10$ & $[0.08, 0.12]$ & $6.48$ & $[6.42, 6.78]$ & $17.02$ & $[16.64, 17.65]$ \\
 & \texttt{RandA1} & $0.01$ & $[0.01, 0.01]$ & $0.71$ & $[0.69, 0.73]$ & $0.56$ & $[0.53, 0.62]$ \\
 & \texttt{RandA2} & $0.08$ & $[0.00, 0.12]$ & $0.75$ & $[0.73, 0.79]$ & $0.94$ & $[0.77, 1.09]$ \\
 & \texttt{RandB} & $0.01$ & $[0.00, 0.02]$ & $0.57$ & $[0.52, 0.61]$ & $0.07$ & $[-0.05, 0.47]$ \\
 & \texttt{MB} & $0.02$ & $[0.01, 0.06]$ & $0.72$ & $[0.70, 0.77]$ & $0.49$ & $[0.43, 0.63]$ \\
 & \texttt{xLOLIB} & $0.12$ & $[0.05, 0.65]$ & $3.87$ & $[3.19, 11.60]$ & $9.81$ & $[7.17, 30.89]$ \\
 & \texttt{xLOLIB2} & $0.34$ & $[0.26, 0.73]$ & $4.18$ & $[3.56, 12.35]$ & $9.88$ & $[7.53, 27.81]$ \\
\cline{2-8}
 & \textbf{Overall} & $0.10$ & $[0.00, 0.73]$ & $3.42$ & $[0.52, 12.35]$ & $7.87$ & $[-0.05, 30.89]$ \\
\bottomrule
\end{tabular}
}
\end{table}

Although these statistics are not strong predictors of instance difficulty, it is apparent that \texttt{EXIOBASE} instances exhibit lower sparsity and variation than \texttt{LOLIB} instances, while the overall skewness is roughly similar.

\section{Multi-Solution Linear Ordering Problem}\label{sec:problem}

We introduce the Multi-Solution LOP (MS-LOP) as a multimodal and diversity-aware extension of the LOP that unifies three related research directions: the Multiple LOP of Benito-Marimom et al.~\cite{benito2025multiple}, and the tasks of multimodal optimization (e.g.,~\cite{preuss2021multimodal}) and evolutionary diversity optimization (e.g.,~\cite{do2022analysis}).

Let $f : \mathbb{S}_n \to \mathbb{R}$ be the LOP objective function defined as in Eq.~\eqref{eq:objfun} and \mbox{$\mathcal{N} : \mathbb{S}_n \to 2^{\mathbb{S}_n}$} be a neighborhood relation on $\mathbb{S}_n$ (e.g., the insertion neighborhood), then $\mathcal{L} = \{ \sigma \in \mathbb{S}_n : f(\sigma) \ge f(\pi), \; \forall \; \pi \in \mathcal{N}(\sigma) \}$ is the set of local optima of $f$ with respect to $\mathcal{N}$. 
Given an integer $m>0$, $\mathcal{L}_m = \{ S \subseteq \mathcal{L} : |S|=m \}$ is the family of sets containing exactly $m$ distinct local optima.

For any $S \in \mathcal{L}_m$, we define its quality $\Phi(S)$ and its diversity $\Delta(S)$.
The quality $\Phi(S)$ is the average of the individual objective values in $S$, while the diversity $\Delta(S)$ can be instantiated by any suitable metric that quantifies the dispersion within $S$ based on the pairwise Kendall's-$\tau$ distances of its members.
The Kendall’s-$\tau$ distance counts the number of pairwise disagreements between two permutations and is widely regarded as the most appropriate distance measure for the LOP~\cite{schiavinotto2004linear,schiavinotto2007review}.
Two reasonable choices for $\Delta$ are: the 
sum of
nearest-neighbor 
distances
$\Delta_\mathit{NN}$~\cite{benito2025multiple,marti2022exact}, and the Solow-Polasky diversity $\Delta_\mathit{SP}$~\cite{preuss2013measuring,solow1994measuring}.
The metrics $\Phi$, $\Delta_\mathit{NN}$ and $\Delta_\mathit{SP}$ are formally defined as:
\begin{equation}
    \Phi(S) = \frac{1}{m} \sum_{\sigma \in S} f(\sigma)
    ,
    \label{eq:phi}
\end{equation}
\begin{equation}
    \Delta_\mathit{NN}(S) = 
    \sum_{\sigma \in S} 
    \min_{\pi \in S \setminus \{\sigma\}} d( \sigma, \pi )
    ,
    \label{eq:delta_nn}
\end{equation}
\begin{equation}
    \Delta_\mathit{SP}(S) = \mathbf{1}_m^T \mathbf{C}^{-1} \mathbf{1}_m
    ,
    \label{eq:delta_sp}
\end{equation}
where: 
$d(\cdot,\cdot)$ is the Kendall's-$\tau$ distance between permutations,
$\mathbf{C} = [c_{ij}]_{m \times m}$ is the similarity matrix with entries $c_{ij} = e^{-\theta d(\sigma_i,\sigma_j) }$ for each pair $\sigma_i,\sigma_j \in S$,
$\theta>0$ is a Solow-Polasky parameter~\cite{solow1994measuring},
and $\mathbf{1}_m$ is a (column) vector of $m$ ones. 



We compare sets in $\mathcal{L}_m$ using a lexicographic order $\succeq$ that prioritizes quality over diversity.
For any $S,P \in \mathcal{L}_m$, we define $S \succeq P$ if and only if either 
\mbox{$\Phi(S)>\Phi(P)$}, or 
$\Phi(S) = \Phi(P)$ and $\Delta(S) \ge \Delta(P)$.
Therefore, the goal of \mbox{MS-LOP} is to identify a set $S^*$ of $m$ distinct local optima that is maximal with respect to $\succeq$.
More formally, 
\begin{equation}
    S^* \in
    \left\{ 
        S \in \mathcal{L}_m : 
            S \succeq P \;\; \forall \,\, P \in \mathcal{L}_m
    \right\}
    \label{eq:goal}
\end{equation}




Interestingly, 
MS-LOP
subsumes several established 
tasks:
the Multiple LOP 
of~\cite{benito2025multiple}, since global optima are also local optima; the objective of multimodal optimization~\cite{preuss2021multimodal}, as quality is prioritized over diversity; and evolutionary diversity optimization~\cite{do2022analysis}, as diversity is explicitly optimized.
Importantly, unlike these approaches, MS-LOP does not require the optimal objective value, which is unavailable in practice due to the NP-hardness of the LOP.

Finally, we suggest addressing the MS-LOP with small set cardinalities, such as $m \in \{5,10,15\}$, since in practical settings the (approximation of the) solution set $S^*$ is ultimately analyzed by a human decision maker.

\section{Algorithms for the Multi-Solution LOP}\label{sec:algs}

As a first approach to the MS-LOP, we propose an algorithmic scheme that can be readily integrated into any single-solution LOP algorithm $\mathcal{A}$ based on local search.
As discussed in Sect.~\ref{sec:lop}, such methods constitute the predominant paradigm in the current state-of-the-art.

The proposed scheme maintains an archive of solutions $S$, initialized as the empty set.
$\mathcal{A}$ is modified in such a way that, during its execution, it triggers the procedure $\mathtt{UpdateArchive}(\sigma)$ whenever a locally optimal solution $\sigma \in \mathbb{S}_n$ is generated.
Upon termination, the archive $S$ contains $m$ distinct local optima that approximate the goal defined in Eq.~\eqref{eq:goal}.

The pseudo-code of $\mathtt{UpdateArchive}(\sigma)$ is shown in Alg.~\ref{alg:ua}.
It
first checks whether $\sigma \in S$.
If it is, no action is taken, since all solutions in $S$ must be distinct.
Otherwise, if $|S|<m$, $\sigma$~is~simply added to the archive.
In the more interesting case where $\sigma \notin S$ and $|S|=m$, a temporary set $S' = S \cup \{\sigma\}$ of $m+1$ solutions is formed.
To determine which solution to exclude, a \mbox{$(m+1)$-dimensional} utility vector $\mathbf{u}^{(\pi)} = (u_0, u_1, \ldots, u_m)$ is constructed for each $\pi \in S'$, where: $u_0 = f(\pi)$ and $u_1 \le u_2 \le \ldots \le u_m$ are the Kendall's-$\tau$ distances between $\pi$ and the remaining members of $S'$, sorted in ascending order.
Finally, the solution $\pi \in S'$ with the lexicographically smallest utility vector $\mathbf{u}^{(\pi)}$ is removed from $S'$ to yield the updated archive $S$.

\begin{algorithm}[t]
	\caption{The algorithm for updating the archive $S$.}
	\begin{algorithmic}[1]
        \Procedure{UpdateArchive}{$\sigma \in \mathbb{S}_n$}
            \If{$\sigma \notin S$}
                \If{$|S|<m$}
                    \State $S \gets S \cup \{\sigma\}$
                \Else
                    \State $S' \gets S \cup \{\sigma\}$
                    \ForAll{$\pi \in S'$}
                        \State $u^{(\pi)}_0 \gets f(\pi)$
                        \State Calculate the $m$ Kendall's-$\tau$ distances $d(\pi,\rho)$ for each $\rho \in S' \setminus \{\pi\}$
                        \State $\langle u^{(\pi)}_1, \ldots, u^{(\pi)}_m \rangle \gets $ sort the distances 
                    \EndFor
                    \State $\pi \gets $ lexicographic argmin of the $\mathbf{u}$ vectors
                    \State $S \gets S' \setminus \{\pi\}$
                \EndIf
            \EndIf
        \EndProcedure
	\end{algorithmic}
	\label{alg:ua}
\end{algorithm}

This procedure, inspired by the selection mechanism of the well-known SPEA2 algorithm~\cite{zitzler2001spea2} widely used in multi-objective optimization, enables the construction of an anytime algorithm, i.e., if interrupted at any iteration, it always returns a solution set of $m$ elements.
The underlying rationale is that $u_0$, i.e., the objective value, serves as the primary selection criterion.
In the case of ties, $u_1$, i.e., the distance to the nearest neighbor, is used as a secondary criterion, and any remaining ties are resolved by recursively reapplying the nearest-neighbor principle.
Consequently, an algorithm $\mathcal{A}$ equipped with $\mathtt{UpdateArchive}$ explicitly maximizes the average objective value while ensuring good diversity in the returned solution set.

Since $m \ll n$, the complexity
of $\mathtt{UpdateArchive}$
is dominated by the computation of the Kendall's-$\tau$ distances, each of which requires $\Theta(n \log n)$ steps.
To improve efficiency, we maintain simple bookkeeping: at each call, only the $m$ distances between the new solution $\sigma$ and those in $S$ are computed.
Hence, the worst-case asymptotic complexity of $\mathtt{UpdateArchive}$ is $\Theta(m n \log n)$.

\section{Experiments}\label{sec:exp}

Experiments were conducted on the proposed \texttt{EXIOBASE} benchmark suite under both single-solution and multiple-solution settings using the two state-of-the-art algorithms CD-RVNS and MA-EDM.
In the single-solution setting, each algorithm was executed 30 times per instance.
For the multiple-solution setting~\mbox{(MS-LOP)}, the algorithms were extended with the scheme introduced in Sect.~\ref{sec:algs}, resulting in the variants MS-CD-RVNS and MS-MA-EDM.
Three archive sizes, $m \in \{5,10,15\}$, were considered, with 30 independent runs per instance and size.
To ensure comparability, we adopt a unified stopping criterion, terminating each run after $100n$ local optima (not necessarily distinct) are generated.
The algorithms were evaluated on the \texttt{rxr}, \texttt{pxp}, and \texttt{os300} instance sets, with the smaller \texttt{isic} instances used in a preliminary experiment to study how the number of global optima varies with instance sparsity.
C\texttt{++} implementations were used, and all experiments were conducted on a machine running Ubuntu 22.04, equipped with an Intel Core i9-11900F CPU (2.50 GHz) and 32 GB of RAM.


\begin{figure}[!b]
    \centering
    \begin{subfigure}[b]{0.30\textwidth}
        \centering
        \includegraphics[width=\textwidth]{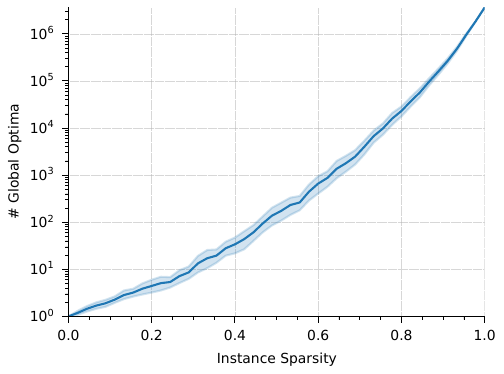}
        \caption{}
        \label{fig:isic_optima}
    \end{subfigure} 
    \hfill
    \begin{subfigure}[b]{0.30\textwidth}
        \centering
        \includegraphics[width=\textwidth]{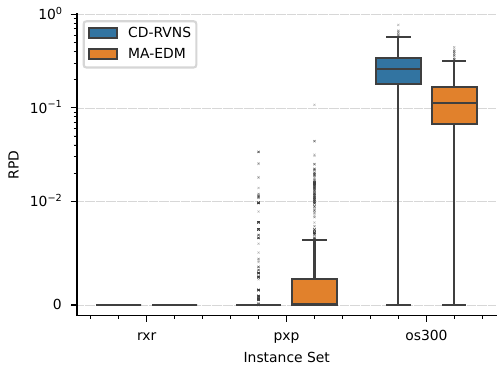}
        \caption{}
        \label{fig:rpd-ss}
    \end{subfigure}
    \hfill
    \begin{subfigure}[b]{0.30\textwidth}
        \centering
        \includegraphics[width=\textwidth]{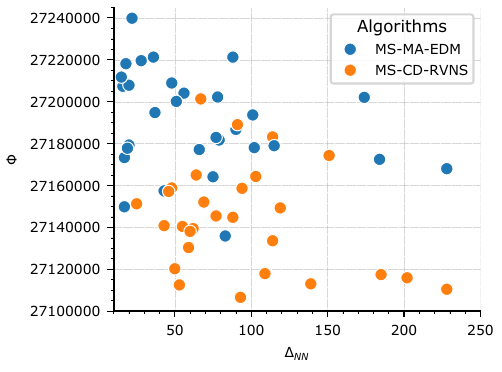}
        \caption{}
        \label{fig:scatter_nn}
    \end{subfigure}
    \caption{\texttt{(a)} Number of global optima (log scale) as a function of sparsity for \texttt{isic} instances. \texttt{(b)} Relative percentage deviations (log scale) in the single-solution setting for the \texttt{rxr}, \texttt{pxp}, and \texttt{os300} instances. \texttt{(c)} $\Phi$ vs. $\Delta_\mathit{NN}$ for all the executions in the \texttt{os300\_it\_2022\_n300} instance with archive size $m=15$.}
\end{figure}

\subsubsection{Number of global optima vs. sparsity.}\label{sec:exp-isic}
A first experiment was conducted to investigate how the number of global optima varies with instance sparsity (percentage of zeros in the LOP matrix). The \texttt{isic} instances were used, as their size ($n=10$) allows exhaustive evaluation of the search space. For each instance, sparsity was gradually increased by zeroing out one additional matrix entry at each iteration, generating a chain of instances ranging from no sparsity to full sparsity. Then, for every generated instance, all $n!$ solutions were evaluated and the number of global optima was recorded. The experiment was repeated for all \texttt{isic} instances, and the results were aggregated in Fig.~\ref{fig:isic_optima}, which reports the mean number of global optima and the corresponding 95\% confidence interval as a function of instance sparsity. The results clearly indicate that the number of global optima increases exponentially as sparsity grows.

\subsubsection{Single-solution experiments.}\label{sec:exp-ss}
To evaluate the effectiveness of CD-RVNS and MA-EDM in the classical LOP setting, we use the standard Relative Percentage Deviation (RPD) metric. 
Let $v$ denote the final objective value obtained from a single execution, and let $\mathit{best}$ be the best objective value observed across all executions of all algorithms on the same instance. 
Then, $v$ is converted to a RPD value as $\mathit{rpd} = 100 \cdot \frac{\mathit{best}-v}{\mathit{best}}$. 
Lower RPDs indicate better effectiveness, and this normalization allows results to be fairly aggregated across instances.

Fig.~\ref{fig:rpd-ss} shows the RPDs obtained from the experiments, aggregated by instance set.
On the smallest \texttt{rxr} instances, both algorithms consistently achieve near-zero RPDs, indicating that all runs converged to essentially the same value.
For medium-sized \texttt{pxp} instances, CD-RVNS is more robust and effective than MA-EDM, while the latter performs better on the larger \texttt{os300} instances.
These observations are supported by per-instance Mann–Whitney U tests at a 0.05 significance level and p-values adjusted via the Benjamini–Hochberg procedure: no significant differences were found on \texttt{rxr}, CD-RVNS significantly outperforms MA-EDM on 29 out of 49 \texttt{pxp} instances while being significantly outperformed on only 4, and MA-EDM is significantly better on 46 out of 49 \texttt{os300} instances.

This behavior is possibly explained by the stronger emphasis of MA-EDM on diversity management, which we believe provides an advantage on larger instances and longer runs, while the trajectory-based strategy of CD-RVNS appears better suited for medium-sized instances.

For detailed results and best known objective values in the classical LOP setting, we refer the interested reader to the supplementary material~\cite{supplementary2026}

\subsubsection{MS-LOP experiments.} \label{sec:exp-ms}
\begin{figure}[b!]
    \centering
    \includegraphics[width=\linewidth]{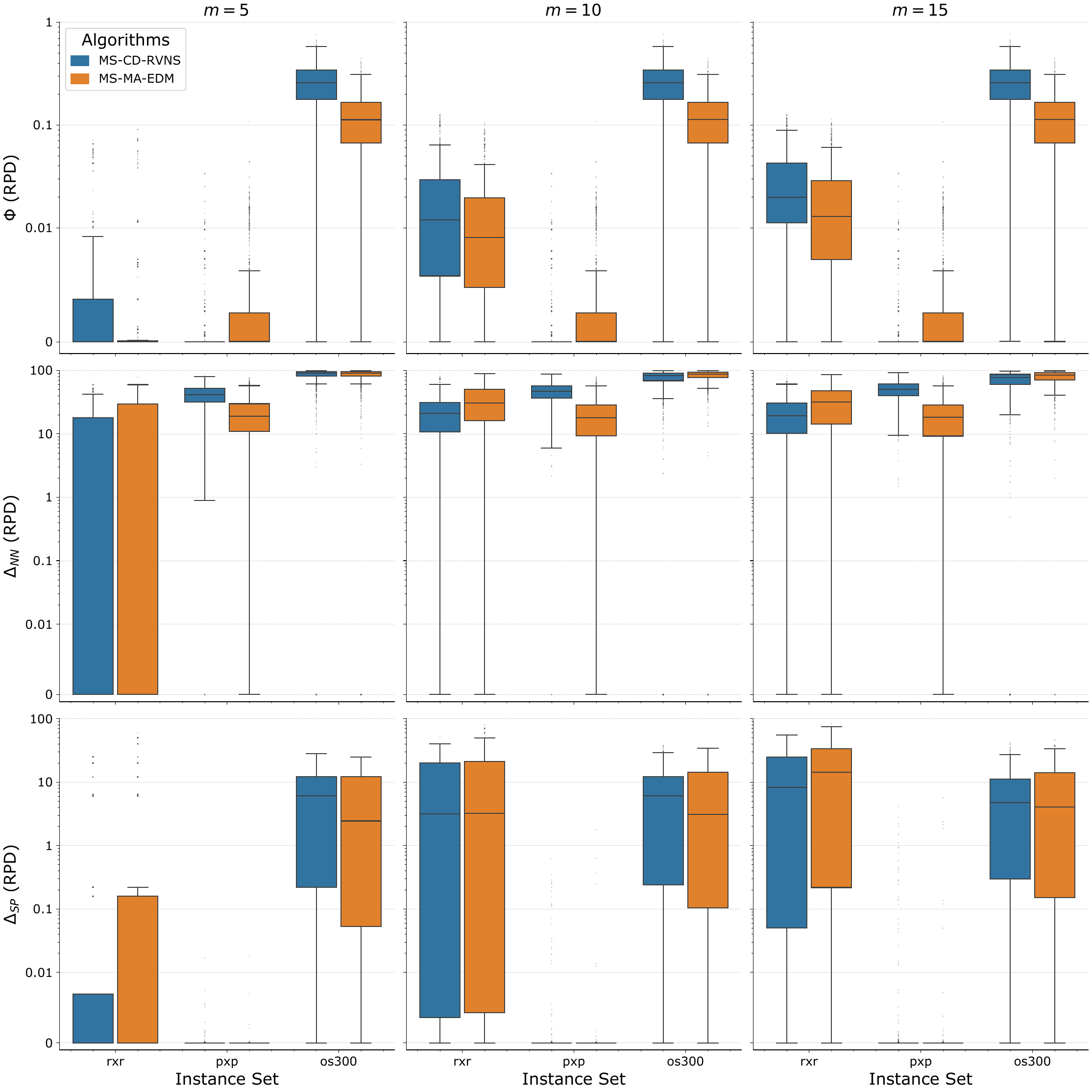}
    \caption{Each row of boxplots shows the relative percentage deviations of the $\Phi$, $\Delta_\mathit{NN}$, and $\Delta_\mathit{SP}$ metrics, respectively. Columns correspond to different archive sizes ($m$ values), and within each plot, the metrics are shown in log scale and aggregated by instance set.}
    \label{fig:boxplot-ms}
\end{figure}

To evaluate the multi-solution variant of the LOP, we assess each run in terms of both solution quality and diversity.
Quality is measured using the $\Phi$ metric as defined in Eq.~\eqref{eq:phi}, while diversity is quantified using $\Delta_\mathit{NN}$ and $\Delta_\mathit{SP}$ as defined in Eqs.~\eqref{eq:delta_nn}~and~\eqref{eq:delta_sp}.
For aggregation and visualization, all measures are normalized using the same RPD technique described earlier (using the best $\Phi$ value for each instance, and the maximum $\Delta_\mathit{NN}$ and $\Delta_\mathit{SP}$ values\footnote{$\Delta_\mathit{SP}$ was computed using normalized Kendall’s-$\tau$ distances, with $\theta$ calibrated such that the median of the observed distances corresponds to a similarity of 0.5.} for each instance-archive~size pair).
The resulting normalized values---which have to be minimized for better quality and diversity, due to the RPD transformation---are presented in log scale as boxplots in Fig.~\ref{fig:boxplot-ms}.

Regarding the $\Phi$ metric, for all $m$ values, similarly to the classical setting, \mbox{MS-CD-RVNS} performs better on \texttt{pxp} instances, while MS-MA-EDM outperforms on the larger \texttt{os300} instances.
Interestingly, the tie observed in the classical LOP for the smaller \texttt{rxr} instances does not occur in the MS-LOP case, where MS-MA-EDM is slightly more effective.
To further investigate this behavior, we analyzed the archive filling patterns and found that while the archive is always filled to its maximum size $m$ for \texttt{pxp} and \texttt{os300} instances, in the \texttt{rxr} case it is often underfilled---likely due to fewer local optima in the instance landscape---, and MS-CD-RVNS surprisingly returns a slightly larger archive on average than MS-MA-EDM (e.g., 10.94 vs. 10.28 for $m=15$), suggesting that its additional solutions may have worse objective values.

Regarding diversity, \texttt{rxr} appears to be the only instance set where diversity depends on the archive size, showing greater diversity for smaller $m$ values.
For \texttt{pxp} instances, the Solow-Polasky diversity remains stable and nearly identical across all runs (approximately 0 deviation) for both algorithms, while the nearest-neighbor diversity of MS-MA-EDM is higher than that of MS-CD-RVNS.
In contrast, for \texttt{os300} instances, there is no clear difference in diversity between the two algorithms.

Finally, in Fig.~\ref{fig:scatter_nn}, for illustrative purposes, we provide a scatter plot showing the relationship between the $\Phi$ quality and the $\Delta_\mathit{NN}$ diversity achieved by each run on the \texttt{os300\_it\_2022\_n300} instance with an archive size of $m=15$.
The figure confirms the previously observation, showing that, on this illustrative instance, MS-MA-EDM achieves higher objective values ($\Phi$) than MS-CD-RVNS, while the distribution of the average Kendall’s-$\tau$ distance to the nearest neighbor within the returned solution set ($\Delta_\mathit{NN}$) is similar for both algorithms.
Furthermore, since each point in the scatter plot of Fig.~\ref{fig:scatter_nn} corresponds to the solution set obtained by a single execution of the algorithms, it becomes evident that neither method consistently returns the same solution set. This variability suggests that there is still room for improvement, particularly in terms of robustness.

For detailed results and statistical analyses in the MS-LOP setting, we refer the interested reader to the supplementary material~\cite{supplementary2026}, which also includes a reproducibility package.
\section{Conclusion and Future Work}\label{sec:concl}

This work has addressed a critical gap in the study of the Linear Ordering Problem (LOP) by modernizing its benchmarking infrastructure and expanding its algorithmic objectives.

By introducing the \texttt{EXIOBASE} benchmark suite, derived from a widely used global multi-regional input-output database, this work provides a contemporary testbed that captures the structural complexity of modern economies from 1995 to 2022.
The newly introduced instances offer a more representative evaluation environment for modern algorithms than traditional benchmarks, which are based on macroeconomic data that are now more than half a century old.
Initial experimental results on the new instances are provided using two recent state-of-the-art LOP algorithms.

Furthermore, this work formalizes the Multi-Solution LOP (MS-LOP), shifting the focus from the search for a single optimum to the identification of diverse sets of high-quality solutions.
The MS-LOP framework unifies three existing approaches to multi-solution optimization: one tailored to the LOP but less suitable for non-exact methods such as metaheuristics, and the two tasks of multimodal optimization and evolutionary diversity optimization that have not previously been applied to the LOP.
The introduction of rigorous metrics for both quality and diversity, alongside the lexicographic prioritization of these objectives, provides a comprehensive framework for future research in multimodal and diversity-aware combinatorial optimization.

To address the MS-LOP, we proposed an algorithmic scheme that can be readily integrated into existing (single-solution) metaheuristics to identify diverse sets of high quality solutions.
Specifically, our approach has been incorporated into CD-RVNS and MA-EDM, two recent state-of-the-art LOP algorithms—one trajectory-based and the other population-based.
Initial experimental results for the MS-LOP are then presented using the proposed methodology and considering the \texttt{EXIOBASE} benchmark suite.

Finally, a promising avenue for future research is the development of algorithms specifically designed for the MS-LOP, rather than adaptations of existing (single-solution) LOP methods.



\bibliographystyle{splncs04}
\bibliography{biblio}

\end{document}